%% file: submission.tex
\documentclass{article}

\usepackage{microtype}
\usepackage{graphicx}
\usepackage{booktabs} 
\usepackage{xspace} 
\usepackage{amsmath} 
\usepackage{comment}
\usepackage{caption}
\usepackage{subcaption}
\usepackage{nicefrac}       
\usepackage{hyperref}
\usepackage{tabularx}

\newcolumntype{C}[1]{>{\centering\arraybackslash}m{#1}}

\newcommand{\alglinelabel}{%
	\addtocounter{ALC@line}{-1}
	\refstepcounter{ALC@line}
	\label
}
\usepackage[accepted]{icml2021}

\icmltitlerunning{Sparse within Sparse Gaussian Processes using Neighbor Information}

\begin{document}
\input{notation}
\twocolumn[
\icmltitle{Sparse within Sparse Gaussian Processes using Neighbor Information}




\begin{icmlauthorlist}
\icmlauthor{Gia-Lac Tran}{eurecom,nus}
\icmlauthor{Dimitrios Milios}{eurecom}
\icmlauthor{Pietro Michiardi}{eurecom}
\icmlauthor{Maurizio Filippone}{eurecom}
\end{icmlauthorlist}

\icmlaffiliation{eurecom}{Department of Data Science, Eurecom, France}
\icmlaffiliation{nus}{Department of Computer Science, National University of Singapore, Singapore}

\icmlcorrespondingauthor{Gia-Lac Tran}{tranlac@nus.edu.sg}

\icmlkeywords{Machine Learning, ICML}

\vskip 0.3in
]



\printAffiliationsAndNotice{} %

\begin{abstract}
\input{abstract}
\end{abstract}

\input{introduction}
\section{Related Work and Background}
\input{related_work}
\input{background}

\input{methods}

\input{experiments}

\input{conclusions}

\paragraph{Acknowledgments}
MF gratefully acknowledges support from the AXA Research Fund and the Agence Nationale de la Recherche (grant ANR-18-CE46-0002 and ANR-19-P3IA-0002).

\newpage

\bibliography{tran}
\bibliographystyle{icml2021}

\end{document}

%% file: notation.tex
\definecolor{mygreen}{rgb}{0.2, 0.7, 0.2}
\definecolor{myorange}{rgb}{0.9, 0.5, 0.0}
\definecolor{mypurple}{rgb}{0.5, 0, 0.5}

\newcommand\noteMF[1]{\textcolor{red}{MF - #1}}
\newcommand\noteGL[1]{\textcolor{blue}{GL - #1}}
\newcommand\notePM[1]{\textcolor{myorange}{PM - #1}}
\newcommand\noteDM[1]{\textcolor{violet}{DM - #1}}
\newcommand\noteEB[1]{\textcolor{mygreen}{EB - #1}}
\newcommand\noteJPC[1]{\textcolor{mypurple}{JPC - #1}}
\newcommand\redtext[1]{\textcolor{red}{#1}}
\newcommand\greentext[1]{\textcolor{green}{#1}}
\newcommand\questiontext[1]{\textcolor{blue}{#1}}

\newcommand{\newpara}{\\\\}
\newcommand{\xvect}{\mathbf{x}} 
\newcommand{\Xcal}{\mathcal{X}}
\newcommand{\noise}{\epsilon} 
\newcommand{\E}{\mathrm{E}} 
\newcommand{\norm}{\mathcal{N}} 
\newcommand{\DD}{\mathcal{D}}
\newcommand{\II}{\mathcal{I}}
\newcommand{\JJ}{\mathcal{J}}
\newcommand{\ZZ}{\mathcal{Z}}
\newcommand{\BB}{\mathcal{B}}
\newcommand{\Brm}{\mathrm{B}}
\newcommand{\ZZvect}{\boldsymbol{\mathcal{Z}}} 
\newcommand{\WW}{\mathcal{W}}
\newcommand{\WWvect}{\boldsymbol{\mathcal{W}}} 
\newcommand{\Xmat}{\mathbf{X}} 
\newcommand{\yvect}{\mathbf{y}} 
\newcommand{\fvect}{\mathbf{f}} 
\newcommand{\zerovect}{\mathbf{0}} 
\newcommand{\onevect}{\mathbf{1}} 
\newcommand{\Kmat}{\mathbf{K}} 
\newcommand{\bigO}{\mathcal{O}} 
\newcommand{\Qmat}{\mathbf{Q}} 
\newcommand{\Nystrom}{\text{Nystr}\ddot{\text{o}}\text{m}} 
\newcommand{\zvect}{\mathbf{z}} 
\newcommand{\muvect}{\boldsymbol{\mu}}
\newcommand{\Amat}{\mathbf{A}}
\newcommand{\Sigmamat}{\boldsymbol{\Sigma}}
\newcommand{\nind}{M}
\newcommand{\nconv}{d} 
\newcommand{\datadim}{D}
\newcommand{\nhidden}{N_{\mathrm{h}}} 
\newcommand{\nclass}{Q} 
\newcommand{\nobs}{N} 
\newcommand{\dimgrid}{P} 
\newcommand{\R}{\mathbb{R}}
\newcommand{\N}{\mathbb{N}}
\newcommand{\Z}{\mathbb{Z}}
\newcommand{\F}{\mathcal{F}}
\newcommand{\I}{\mathcal{I}}
\newcommand{\LL}{\mathcal{L}}
\newcommand{\uu}{\mathbf{u}}
\newcommand{\ee}{\mathbf{e}}

\newcommand{\KL}{\mathrm{KL}}
\newcommand{\ELL}{\mathrm{ELL}}
\newcommand{\const}{\mathrm{const.}}
\newcommand{\diag}{\mathrm{diag}}
\newcommand{\Tr}{\mathrm{Tr}}

\newcommand{\avect}{\mathbf{a}}
\newcommand{\cvect}{\mathbf{c}}
\newcommand{\dvect}{\mathbf{d}}
\newcommand{\evect}{\mathbf{e}}
\newcommand{\gvect}{\mathbf{g}}
\newcommand{\hvect}{\mathbf{h}}
\newcommand{\mvect}{\mathbf{m}}
\newcommand{\pvect}{\mathbf{p}}
\newcommand{\svect}{\mathbf{s}}
\newcommand{\rvect}{\mathbf{r}}
\newcommand{\uvect}{\mathbf{u}}
\newcommand{\vvect}{\mathbf{v}}
\newcommand{\wvect}{\mathbf{w}}
\newcommand{\tvect}{\mathbf{t}}
\newcommand{\onesvect}{\mathbf{1}}

\newcommand{\Bmat}{\mathbf{B}}
\newcommand{\Cmat}{\mathbf{C}}
\newcommand{\Dmat}{\mathbf{D}}
\newcommand{\Zmat}{\mathbf{Z}}
\newcommand{\Ymat}{\mathbf{Y}}
\newcommand{\Fmat}{\mathbf{F}}
\newcommand{\Wmat}{\mathbf{W}}
\newcommand{\Gmat}{\mathbf{G}}
\newcommand{\Hmat}{\mathbf{H}}
\newcommand{\Jmat}{\mathbf{J}}
\newcommand{\Imat}{\mathbf{I}}
\newcommand{\Pmat}{\mathbf{P}}
\newcommand{\Smat}{\mathbf{S}}
\newcommand{\Lmat}{\mathbf{L}}
\newcommand{\Mmat}{\mathbf{M}}
\newcommand{\Umat}{\mathbf{U}}
\newcommand{\Vmat}{\mathbf{V}}
\newcommand{\Phimat}{\boldsymbol{\Phi}}
\newcommand{\Pimat}{\boldsymbol{\Pi}}
\newcommand{\Omegamat}{\boldsymbol{\Omega}}
\newcommand{\Psimat}{\boldsymbol{\Psi}}
\newcommand{\Ximat}{\boldsymbol{\Xi}}
\newcommand{\Thetamat}{\boldsymbol{\Theta}}
\newcommand{\Lambdamat}{\boldsymbol{\Lambda}}

\newcommand{\alphavect}{\boldsymbol{\alpha}}
\newcommand{\betavect}{\boldsymbol{\beta}}
\newcommand{\thetavect}{\boldsymbol{\theta}} 
\newcommand{\psivect}{\boldsymbol{\psi}}
\newcommand{\etavect}{\boldsymbol{\eta}}
\newcommand{\rhovect}{\boldsymbol{\rho}}
\newcommand{\tauvect}{\boldsymbol{\tau}}
\newcommand{\nuvect}{\boldsymbol{\nu}}
\newcommand{\omegavect}{\boldsymbol{\omega}}
\newcommand{\sigmavect}{\boldsymbol{\sigma}}
\newcommand{\zetavect}{\boldsymbol{\zeta}}
\newcommand{\varepsilonvect}{\boldsymbol{\epsilon}}
\newcommand{\deltavect}{\boldsymbol{\delta}}
\newcommand{\varphivect}{\boldsymbol{\varphi}}
\newcommand{\phivect}{\boldsymbol{\phi}}
\newcommand{\xivect}{\boldsymbol{\xi}}

\newcommand{\name}[1]{{\textsc{#1}}\xspace}

\newcommand{\mcmc}{\name{mcmc}}


\newcommand{\cifar}{\name{cifar10}}
\newcommand{\mnisteight}{\textsc{mnist}8\textsc{m}\xspace}
\newcommand{\onedimension}{\name{1d}}
\newcommand{\protein}{\name{protein}}
\newcommand{\powerplant}{\name{powerplant}}
\newcommand{\kin}{\name{kin}}
\newcommand{\naval}{\name{naval}}
\newcommand{\spam}{\name{spam}}
\newcommand{\eeg}{\name{eeg}}
\newcommand{\credit}{\name{credit}}
\newcommand{\banana}{\name{banana}}
\newcommand{\usps}{\name{usps}}
\newcommand{\cleveland}{\name{cleveland}}
\newcommand{\banknote}{\name{banknote}}
\newcommand{\wisconsin}{\name{wisconsin}}
\newcommand{\wave}{\name{wave}}
\newcommand{\query}{\name{query}}
\newcommand{\uci}{\name{uci}}
\newcommand{\mnist}{\name{mnist}} 
\newcommand{\notmnist}{\name{not-mnist}} 
\newcommand{\cifart}{\name{cifar10}}
\newcommand{\cifarh}{\name{cifar100}}
\newcommand{\airline}{\name{airline}}
\newcommand{\irradiance}{\name{irradiance}}

\newcommand{\autogp}{\textsc{a}uto\textsc{gp}\xspace}
\newcommand{\gpflow}{\textsc{gpf}low\xspace}
\newcommand{\tensorflow}{\textsc{T}ensor\textsc{F}low\xspace}

\newcommand{\arccosine}{\name{arc-cosine}}
\newcommand{\dgprbf}{\textsc{dgp}-\textsc{rbf}\xspace}
\newcommand{\dgparc}{\textsc{dgp}-\textsc{arc}\xspace}
\newcommand{\dgpep}{\textsc{dgp}-\textsc{ep}\xspace}
\newcommand{\dgpvar}{\textsc{dgp}-\textsc{var}\xspace}

\newcommand{\elbo}{\name{elbo}}
\newcommand{\kiss}{\textsc{kiss-gp}}

\newcommand{\gp}{\name{gp}}
\newcommand{\samp}{\name{samp}}
\newcommand{\samppost}{\name{samp-post}}
\newcommand{\gps}{\textsc{gp}s\xspace}
\newcommand{\dgp}{\name{dgp}}
\newcommand{\dgps}{\textsc{dgp}s\xspace}
\newcommand{\dnn}{\name{dnn}}
\newcommand{\dnns}{\textsc{dnn}s\xspace}
\newcommand{\svm}{\name{svm}}
\newcommand{\svms}{\textsc{svm}s\xspace}
\newcommand{\vargp}{\textsc{var}-\textsc{gp}\xspace}
\newcommand{\svgp}{\textsc{svgp}\xspace}
\newcommand{\svgps}{\textsc{svgp}s\xspace}
\newcommand{\sisgp}{\textsc{sisgp}\xspace}
\newcommand{\nngp}{\textsc{nngp}\xspace}

\newcommand{\convweights}{\Psimat}
\newcommand{\ard}{\name{ard}}
\newcommand{\relu}{{\textsc{r}}e\name{lu}}
\newcommand{\arc}{\name{arc}}
\newcommand{\rbf}{\name{rbf}}
\newcommand{\linear}{\name{linear}}
\newcommand{\polynomial}{\name{polynomial-3}}
\newcommand{\gpu}{\name{gpu}}
\newcommand{\opu}{\name{opu}}
\newcommand{\mcd}{\name{mcd}}
\newcommand{\adf}{\name{adf}}
\newcommand{\ivm}{\name{ivm}}
\newcommand{\bcnn}{\name{bcnn}}
\newcommand{\idea}{\name{swsgp}}
\newcommand{\swsgp}{\name{swsgp}}
\newcommand{\swsgpu}{\name{swsgp-u}}
\newcommand{\kissgp}{\name{kiss-gp}}
\newcommand{\ski}{\name{ski}}

\newcommand{\Mip}{\name{m}}
\newcommand{\Hip}{\name{h}}

\newcommand{\mnll}{\name{mnll}}
\newcommand{\rmse}{\name{rmse}}
\newcommand{\nelbo}{\name{nelbo}}
\newcommand{\err}{\name{err}}
\newcommand{\brier}{\name{brier}}
\newcommand{\ece}{\name{ece}}
\newcommand{\entropy}{\name{h}}

\newcommand{\lenet}{\name{LeNet}}
\newcommand{\resnet}{\name{resnet}}

\newcommand{\cnn}{\name{cnn}}
\newcommand{\cnns}{\textsc{cnn}s\xspace}
\newcommand{\gpdnn}{\name{gpdnn}}
\newcommand{\cgp}{\name{cgp}}
\newcommand{\svdkl}{\name{svdkl}}
\newcommand{\temp}{\name{temp}}

\newcommand{\sorf}{\name{sorf}}
\newcommand{\cnngprf}{\name{cnn+gp(rf)}}
\newcommand{\cnngpsorf}{\name{cnn+gp(sorf)}}
\newcommand{\cnngprfmcd}{\name{cnn+gp(rf)+mcd}}

\newcommand\argmin[1]{\underset{#1}{\mathrm{argmin}}}
\newcommand\argmax[1]{\underset{#1}{\mathrm{argmax}}}

%% file: abstract.tex
Approximations to Gaussian processes (GPs) based on inducing variables, combined with variational inference techniques, enable state-of-the-art sparse approaches to infer GPs at scale through mini-batch based learning. 
In this work, we further push the limits of scalability of sparse GPs by allowing large number of inducing variables without imposing a special structure on the inducing inputs.
In particular, we introduce a novel hierarchical prior, which imposes sparsity on the set of inducing variables. We treat our model variationally, and we experimentally show considerable computational gains compared to standard sparse GPs when sparsity on the inducing variables is realized considering the nearest inducing inputs of a random mini-batch of the data. We perform an extensive experimental validation that demonstrates the effectiveness of our approach compared to the state-of-the-art. Our approach enables the possibility to use sparse GPs using a large number of inducing points without incurring a prohibitive computational cost.



%% file: introduction.tex


\section{Introduction}
Gaussian Processes (\gps) \citep{Rasmussen06} offer a powerful framework to perform inference over functions; being Bayesian, \gps provide rigorous uncertainty quantification and prevent overfitting.
However, the applicability of \gps on large datasets is hindered by their computational complexity of $\bigO\left(N^3\right)$, where $N$ is the training size.
This issue has fuelled a considerable amount of research towards scalable \gp methodologies that operate on a set of \emph{inducing variables} \citep{Candela05}.
In the literature, there is a plethora of approaches that offer different treatments of the inducing variables \citep{Lawrence02, Seeger03fastforward, Snelson05, Guzman07, Titsias09, Hensman13, Wilson15, Hensman15b}.
Some of the more recent approaches, such as Scalable Variational Gaussian Processes (\svgps) \citep{Hensman15b}, allow for the application of \gps to problems with millions of data-points. 
In most applications of scalable \gps, these are approximated using $\nind$ inducing points, which results in a complexity of $\bigO\left(M^3\right)$.
It has been shown recently by \citet{Burt19a} that it is possible to obtain an arbitrarily good approximation for a certain class of \gp models (i.e.\ conjugate likelihoods, concentrated distribution for the training data) with $M$ growing more slowly than $N$. However, the general case remains elusive and it is still possible that the required value for $M$ may exceed a certain computational budget.
Our result contributes to strengthen our belief that sparsity does not only enjoy desirable theoretical properties, but it also constitutes an extremely computationally efficient method in practice.

In this work, we push the limits of scalability and effectiveness of sparse \gps enabling a further reduction in complexity, which can be translated to higher accuracy by considering a larger set of inducing variables.
The idea is to operate on a subset of $H$ inducing points during training and prediction, with $H \ll \nind$, while maintaining a sparse approximation with $\nind$ inducing variables.
We formalize our strategy by imposing a sparsity-inducing structure on the prior over the inducing variables and by carrying out a variational formulation of this model. 
This extends the original \svgp framework and enables mini-batch based optimization for the variational objective.
We then consider ways to select the set of $H$ inducing points based on neighbor information; at training time, for a given mini-batch, we activate $H$ out of $M$ inducing variables considering the nearest inducing inputs to the samples in the mini-batch, whereas at test time we select inducing variables corresponding to the inducing inputs which are nearest to the test data-points.
We name our proposal \emph{Sparse within a Sparse \gp} (\idea).
\idea is characterized by a number of attractive features: (i) it improves significantly the prediction quality using a small number of neighboring inducing inputs, and (ii) it accelerates the training phase, especially when the total number of inducing points becomes large. 
We extensively validate these properties on a variety of regression and classification tasks.
We also showcase \idea on a large scale classification problem with $M = 100,000$; we are not aware of other approaches that can handle such a large set of inducing inputs without imposing some special structure on them (e.g., grid) or without considering one-dimensional inputs. 

Hierarchical priors are often applied in Bayesian modeling to achieve compression and to improve flexibility \citep{Molchanov17, Louizos17}.
To the best of our knowledge, this work is the first to explore these ideas as means to sparsify the inducing set in sparse \gps .

%% file: related_work.tex
Sparse \gps that operate on inducing inputs have been extensively studied in the last 20 years \citep{Csato02,Lawrence02,Snelson05,Candela05,Guzman07}.
Many attempts on sparse \gps specified inducing inputs by satisfying certain criteria that produce an informative set of inducing variables \citep{Csato02,Lawrence02,Seeger03fastforward}.
A different treatment has been proposed by \citet{Titsias09}, which involves formulating the selection of inducing inputs as optimization of a variational lower bound to the marginal likelihood.
The variational framework was later extended so that stochastic optimization can be admitted, thus improving scalability for regression \citep{Hensman13} and classification \citep{Hensman15b}, and to deal with large-dimensional inputs \citep{Panos2018}.
All the aforementioned methodologies share a computational complexity of $\bigO\left(\nind^3\right)$.
Although there have been some attempts in the literature to infer the appropriate number of inducing points as well as the inducing inputs \citep{Pourhabib2014,Burt19a}, 
a large number of inducing variables is desirable to improve posterior approximation.
In this work, we present a methodology that builds on the \svgp framework \citep{Hensman15b} and reduces its complexity, thus increasing the potential of sparse \gp application on even larger datasets and with a larger set of inducing variables.

A different approach to scalable \gps was introduced by \citet{Wilson15}, namely Kernel Interpolation for Scalable Structured \gps (\kissgp). 
This line of work involves arranging a large number of inducing inputs into a grid structure; this allows one to scale to very large datasets by means of fast linear algebra. 
The applicability of \kissgp on higher-dimensional problems has been addressed by \citet{Wilson15a} by means of low-dimensional projections.
A more recent extension allows for a constant-time variance prediction using Lanczos methods \citep{Pleiss2018}.
Our work takes a different approach by keeping the \gp prior intact, and by imposing sparsity on the set of inducing variables.

The local approximation of \gps inspired by the the concept of divide-and-conquer is also a practical solution to implement scalable \gps \citep{Kim05, Urtasun08, Datta16, Park16,Park17}. 
More recently, \citet{liu19c} proposed an amortized variational inference framework where the nearest training points are selected for inference.
In our work, we use neighbor information in a different way, by incorporating it in a certain hierarchical structure of the auxiliary variables through a variational scheme.

%% file: background.tex
\subsection{Scalable Variational Gaussian Processes}
\label{subsection: svgp}
Consider a supervised learning problem with inputs $\Xmat = (\xvect_1, \ldots, \xvect_N)^{\top}$ associated with labels $\yvect = ( y_1, \ldots, y_N )^{\top}$. 
Given a set of latent variables $\fvect = (f_1, \ldots, f_N)^{\top}$, \gp models assume that labels are stochastic realizations based on $\fvect$ and a likelihood function $p(\yvect \mid \fvect)$. 
In \svgps, the set of inducing points is characterized by inducing inputs $\Zmat = (\zvect_1, \ldots, \zvect_{\nind})^{\top}$ and inducing variables $\uvect = (u_1, \ldots, u_{\nind})^{\top}$.
Regarding $\fvect$ and $\uvect$, we have the following joint prior:
\begin{equation}
	p(\fvect, \uvect) = \norm\left(0, 
	\begin{bmatrix}
	\Kmat_\Xmat 	& \Kmat_{\Xmat, \Zmat} \\
	\Kmat_{\Zmat, \Xmat} 	& \Kmat_\Zmat
	\end{bmatrix}
	\right) \text{,}
\label{eq:fu_joint}
\end{equation}
where $\Kmat_\Xmat$, $\Kmat_\Zmat$ and $\Kmat_{\Xmat, \Zmat}$ are covariance matrices evaluated at the inputs indicated by the subscripts.
The posterior over inducing variables 
is approximated by a variational distribution $q\left(\uvect\right)=\norm\left(\uvect \mid \mvect, \Smat\right)$, while keeping the exact conditional $p(\fvect \mid \uvect)$ intact, that is $q(\fvect, \uvect) = p(\fvect \mid \uvect) q(\uvect)$.
The variational parameters $\mvect$ and $\Smat$, as well as the inputs $\Zmat$, are optimized by maximizing a lower bound on the marginal likelihood $p(\yvect \mid \Xmat) = \int p(\yvect \mid \fvect) p(\fvect \mid \Xmat) d\fvect$. 
The lower bound on $\log p\left(\yvect \mid \Xmat \right)$ can be obtained by considering the form of $q(\fvect, \uvect)$ above and by applying Jensen's inequality:
\begin{equation}
\label{eq: lower bound on marginal likelihood 1}
\begin{split}
\E_{q\left(\fvect\right)}\left[\log p\left(\yvect \mid \fvect\right)\right] - \KL\left(q\left(\uvect\right) \| \, p\left(\uvect\right)\right) \text{.}
\end{split}
\end{equation}
The approximate posterior $q\left(\fvect\right)$ can be computed by integrating out $\uvect$:
$
q\left(\fvect\right) = \int q\left(\uvect\right) p\left(\fvect\left|\uvect\right.\right) d\uvect.
$
Thanks to the Gaussian form of $q\left(\uvect\right)$, $q\left(\fvect\right)$ can be computed analytically:
\begin{equation}
\label{eq: q(f)}
q\left(\fvect\right) = \norm\left(\fvect\ |\  \Amat\mvect, \ \Kmat_\Xmat + \Amat\left(\Smat - \Kmat_\Zmat\right)\Amat\right) \text{,}
\end{equation}
where $\Amat = \Kmat_{\Xmat, \Zmat} \Kmat_{\Zmat}^{-1}$.
When the likelihood factorizes over training points, 
the lower bound can be re-written as:
\begin{equation}
\textstyle\sum_{i=1}^N\E_{q\left(f_i\right)}\left[\log p\left(y_i\ |\ f_i\right)\right] - \KL\left(q\left(\uvect\right) \| \, p\left(\uvect\right)\right) \text{.}
\end{equation}
Each term of the one-dimensional expectation of the log-likelihood can be computed by Gauss-Hermite quadrature for any likelihoods (and analytically for the Gaussian likelihood).
The $\KL\left(q\left(\uvect\right)\ \|\ p\left(\uvect\right)\right)$ term can be computed analytically given that $q\left(\uvect\right)$ and $p\left(\uvect\right)$ are both Gaussian.
To maintain positive-definiteness of $\Smat$ and perform unconstrained optimization, $\Smat$ is parametrized as $\Smat=\Lmat\Lmat^T$, with $\Lmat$ lower triangular. 

%% file: methods.tex
\section{Sparse Within Sparse GP
}
We present a novel formulation of sparse \gps, which permits the use of a random subset of the inducing points	 with little loss in performance.
We introduce a set of binary random variables $\wvect \in \{0, 1\}^{\nind}$ to govern the inclusion of inducing inputs $\Zmat$ and the corresponding variables $\uvect$. 
We then employ these random variables to define a hierarchical structure on the prior as follows: 
\begin{equation}
\label{eq: hierarchical prior}
p\left(\uvect \mid \wvect \right) = \norm\left(\zerovect, \Dmat_\wvect\Kmat_{\Zmat}\Dmat_\wvect \right),
\end{equation}
where $\Dmat_\wvect = \diag\left(\wvect\right)\text{, and }\wvect \sim p\left(\wvect\right)$.
Although the marginalized prior $p(\uvect)$ is not Gaussian, it is possible to use the joint $p(\uvect, \wvect) = p(\uvect\mid\wvect)\ p(\wvect)$ within a variational scheme.
We thus consider a random subset of the inducing points during the evaluation of the prior in the variational scheme that follows; no inducing points are permanently removed.
Regarding $p(\wvect)$, we consider an implicit distribution: its analytical form is unknown, but we can draw samples from it.
Later, we will consider $p(\wvect)$ based on the nearest inducing inputs to random mini-batches of data.

%

\paragraph{Remarks on the prior over $\fvect$}
Our strategy simply assumes a certain structure on the auxiliary variables, but it \textit{has no effect} on the prior over $\fvect$; the latter remains unchanged.
Let $\II$ and $\JJ$ bet the sets of indices such that $\wvect_{\II} = \onevect$ and $\wvect_{\JJ} = \zerovect$.
Given an appropriate ordering, the conditional $\uvect\mid\wvect$ is effectively the element-wise product $[\uvect_{\II}, \uvect_{\JJ}]^\top = \uvect \circ \wvect$.
This reduces the variances and covariances of some elements of $\uvect$ to zero yielding a distribution of this form:
\begin{equation}
p\left(\fvect, \uvect\mid\wvect\right) =
\norm\left(
\begin{bmatrix}
\zerovect\\
\zerovect\\
\zerovect
\end{bmatrix}
,
\begin{bmatrix}
\Kmat_{\Xmat} & \Kmat_{\Xmat, \Zmat_{\II}} & \zerovect\\
\Kmat_{\Zmat_{\II}, \Xmat} & \Kmat_{\Zmat_{\II}} & \zerovect\\
\zerovect & \zerovect & \zerovect 
\end{bmatrix}
\right)
\end{equation}
The rows and columns of $\uvect_{\JJ}$ can simply be ignored.
Regardless of the value of $\wvect$, the conditional $\fvect,\uvect_{\II}\mid\wvect$ is always a Gaussian marginal, as it is a subset of Gaussian variables.
The marginalized $p(\fvect,\uvect) = \int p(\fvect,\uvect\mid\wvect)\, p(\wvect) d\wvect$ is mixture of Gaussian densities, where the marginal over $\fvect$ is the same for every component of the mixture.

\begin{figure}
\includegraphics[width=\linewidth]{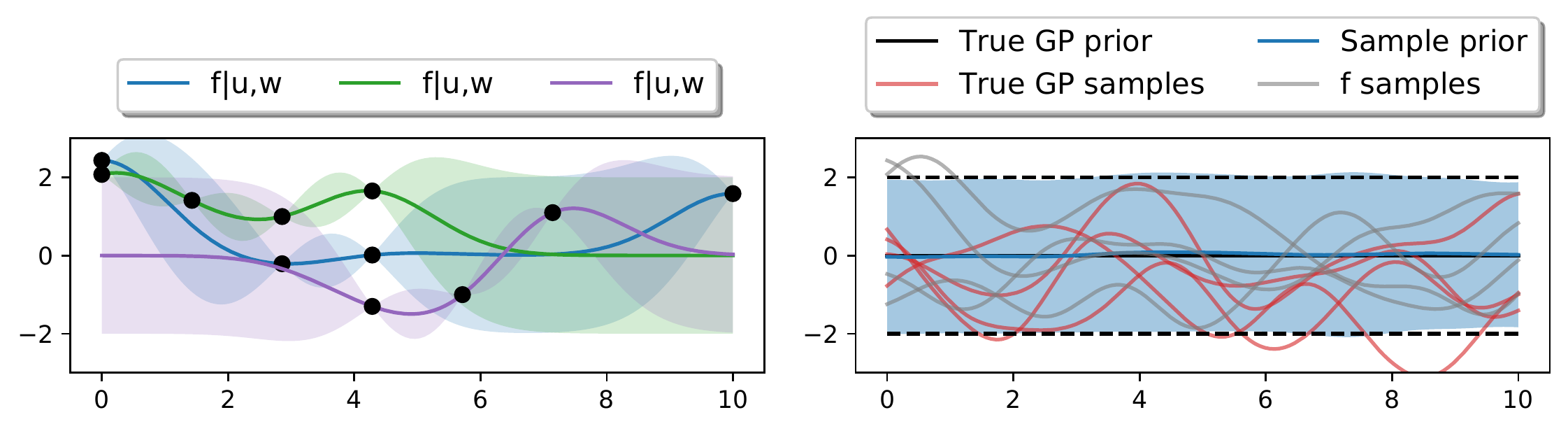}  
\caption{The choice of inducing points does not affect the prior samples drawn from $p(\fvect)$. Left: visualizations of $\fvect\mid\uvect,\wvect$ for different samples of $\wvect$. Right: comparison of the marginalized (w.r.t.~$\uvect,\wvect$) prior over $\fvect$, against the true $p(\fvect)$.}
\label{fig:prior_f}
\end{figure}

The effect on $\fvect$ is demonstrated in Figure \ref{fig:prior_f}, where we sample from the (non-Gaussian) marginalized prior $p(\uvect)$ in two steps: first we consider an arbitrary random subset $\uvect_{\II}$, and then we sample from $p(\uvect_{\II}) \equiv p(\uvect\mid\wvect)$.
Finally, $\fvect$ samples are drawn from $p(\fvect\mid\uvect_{\II})$, which only involves the selected inducing variables $\uvect_{\II}$.
Following Eq.~\eqref{eq:fu_joint}, the conditional $\fvect\mid\uvect$ is normally-distributed with mean
$\mvect_{\fvect\mid\uvect_{\II}} = \Kmat_{\Xmat, \Zmat_{\II}} \Kmat_{\Zmat_{\II}}^{-1} \uvect_{\II}$ 
and covariance  
$\Smat_{\fvect\mid\uvect_{\II}} = \Kmat_\Xmat - \Kmat_{\Xmat, \Zmat_{\II}} \Kmat_{\Zmat_{\II}}^{-1} \Kmat_{\Zmat_{\II}, \Xmat}$.
These conditionals can be seen for different samples of $\uvect, \wvect$ in the left side of Figure \ref{fig:prior_f}, while in the right side we compare the marginalized prior over $\fvect$ against the true \gp prior.

Of course, although the prior remains unchanged, that is not the case for the posterior approximation.
It is well known that the choice of inducing inputs has an effect on the variational posterior \citep{Titsias09,Burt19a}.
Our choice to impose a hierarchical structure to the inducing variables through $\wvect$ effectively changes the model compared to \svgp, and we adapt the variational scheme accordingly.

\subsection{Lower Bound on Marginal Likelihood} 
By introducing $\uvect, \wvect$ and using Jensen's inequality, the lower bound on $\log p\left(\yvect\right)$ can be obtained as follows
\begin{equation}
\label{eq:Jensen1}
\begin{split}
\E_{q\left(\uvect,\wvect\right)} \log p\left(\yvect\left|\uvect,\wvect\right.\right) - \KL\left(q\left(\uvect,\wvect\right) \| p\left(\uvect,\wvect\right)\right),
\end{split}
\end{equation}
where we choose the variational distribution $q$ to reflect the hierarchical structure of the prior, i.e.\  $q\left(\uvect,\wvect\right) = q(\uvect \mid \wvect)\ p(\wvect)$.
This choice enforces sparsity over the approximate posterior $q$; the variational parameters are shared among the conditionals $q(\uvect \mid \wvect)$, for which we assume:
\begin{equation}
q(\uvect \mid \wvect) = \norm\left(\uvect\left|\Dmat_\wvect \mvect, \Dmat_\wvect \Smat \Dmat_\wvect\right.\right)
\label{eq: variational distribution q(u|w)}
\end{equation}
By maximizing the variational bound, we aim to obtain a $q$ that performs well under a sparsified inducing set.
We continue by applying Jensen's inequality on $p\left(\yvect\left|\uvect,\wvect\right.\right)$, obtaining:
\begin{equation}
\label{eq:Jensen2}
\log p\left(\yvect\left|\uvect,\wvect\right.\right) \geq \E_{p\left(\fvect|\uvect,\wvect\right)}\log p\left(\yvect\left|\fvect\right.\right)
\end{equation}
The bound we describe in Equations \eqref{eq:Jensen1} and \eqref{eq:Jensen2} is the same as in \svgp, but with a different variational distribution.
In our case, the variational distribution imposes sparsity for $\uvect$ by means of $\wvect$.
We can now substitute \eqref{eq:Jensen2} into \eqref{eq:Jensen1}, obtaining a bound where
we expand $q(\uvect, \wvect)$
as $q\left(\uvect\left|\wvect\right.\right) p\left(\wvect\right)$.
By making this assumption, we get the following evidence lower bound $\LL_\elbo$:
\begin{equation}
\label{eq:ll_elbo}
	\begin{split}
		\sum_{n=1}^\nobs \E_{p\left(\wvect\right)} \bigg[&\E_{q\left(\uvect\left|\wvect\right.\right)} \E_{p\left(f_n\left|\uvect,\wvect\right.\right)} \log p\left(y_n\left|f_n\right.\right) \\
		&- \frac{1}{N}\KL\left(q\left(\uvect\left|\wvect\right.\right) \bigg\| \, p\left(\uvect\left|\wvect\right.\right)\right)\bigg]
	\end{split}
\end{equation}
Recall that $p\left(\wvect\right)$ is implicit: although we do not make any particular assumptions about its analytical form, we can draw samples from it.
Using MC sampling from $p\left(\wvect\right)$, we can obtain the approximation $\tilde{\mathcal{L}}_\elbo$:
\begin{equation}
	\begin{split}
		\sum_{n=1}^N \bigg[&\E_{q\left(\uvect\left|\tilde{\wvect}^{(n)}\right.\right)} \E_{p\left(f_n\left|\uvect, \tilde{\wvect}^{(n)}\right.\right)} \log p\left(y_n\left|f_n\right.\right) \\
		&- \frac{1}{N}\KL\left(q\left(\uvect\left|\tilde{\wvect}^{(n)}\right.\right) \bigg\| \, p\left(\uvect\left|\tilde{\wvect}^{(n)}\right.\right)\right)\bigg],
	\end{split}
\end{equation}
where $\tilde{\wvect}^{(n)}$ is sampled from $p\left(\wvect\right)$.
\paragraph{Sampling from the set of inducing points.}
Recall that any sample $\tilde{\wvect}$ from $p\left(\wvect\right)$ is a binary vector, i.e. $\wvect \in \{0, 1\}^\nind$.
In case all elements of $\wvect$ are set to one, our approach recovers the original \svgp with computational cost 
of $\bigO\left(\nind^3\right)$ coming from computing $p\left(f_n\left|\uvect,\tilde{\wvect}=\onevect\right.\right)$ and $\mathrm{KL}\left(q\left(\uvect\left|\wvect\right.\right) \| p\left(\uvect\left|\wvect\right.\right)\right)$ in the \elbo.
When any $\tilde{w}_i$ is set to zero, the entries of the $i$-th row and $i$-th column of the covariance matrix in $p\left(\uvect\left|\wvect\right.\right)$ and $q\left(\uvect\left|\wvect\right.\right)$ are zero.
This means that the $i$-th variable becomes unnecessary, so we get rid of $i$-th row and column in these matrices, and also eliminate the $i$-th element in mean vectors of $q\left(\uvect\left|\wvect\right.\right)$ and $p\left(\uvect\left|\wvect\right.\right)$.
This is equivalent to selecting a set of active inducing points in each training iteration. 

\subsection{H-nearest Inducing Inputs}
Despite the fact that $p(\wvect)$ is an implicit distribution, 
we have been able to define and calculate a variational bound, assuming we can sample from $p(\wvect)$. 
We shall now describe our sampling strategy, which relies on neighbor information of random mini-batches.

We introduce $\Zmat^H_\xvect$ as the set of $H$-nearest inducing inputs. 
Intuitively, the prediction for an unseen data $\xvect$ using $\Zmat^H_\xvect$ is a good approximation of the prediction using all $\nind$ inducing points, that is $\Zmat^\nind_\xvect$. 
This can be verified by looking at the predictive mean, which is expressed as a linear combination of kernel functions evaluated between training points and a test point, as in Eq.~\eqref{eq: q(f)}.
The majority of the contribution is given by the inducing points with the largest kernel values, so we can use this as a criterion to establish whether an inducing input is ``close'' to an input vector
(the effect of different kernels on the definition of nearest neighbors is explored in the supplement).
With this intuition, $p\left(\wvect\right)$ becomes a deterministic function $w\left(\xvect\right)$ indicating which inducing inputs are activated. 
For mini-batch based training, the value of $\wvect$ remains random, as it depends on the elements $\xvect$ that are selected in the random mini-batch; this materializes the sampling from the implicit distribution $p(\wvect)$.
The maximization of the \elbo in the setting described is summarized in Algorithm~\ref{alg: MAIP} (\idea).
\begin{algorithm}
	\caption{Sparse within sparse \gp  (\idea)}
	\label{alg: MAIP}
	\begin{algorithmic}[1]
		\REQUIRE $\DD$, $H$, $M$.
		\ENSURE The optimum of trainable parameters $\thetavect$.
		\STATE \alglinelabel{alg: initialize parameters} Initialize $\thetavect$ including kernel's parameters, $\Zmat$, $\mvect$ and $\Lmat$ which can be used to construct $\Smat$, i.e. $\Smat = \Lmat\Lmat^T$.
		\WHILE{stopping criteria is False}\label{alg: while}
			\STATE \alglinelabel{alg: initialize ell and kl} $\ELL \gets 0$ and $\KL \gets 0$.
			\STATE \alglinelabel{alg: extract minibatch-size} Sample mini-batch $\Brm$ of size $n_{\Brm}$ from $\DD$.
			\FOR{$\left(\xvect_i, y_i\right) \in \Brm$} \alglinelabel{alg: for}
				\STATE \alglinelabel{alg: find neighbor inducing points} Find $\Zmat^H_{\xvect_i}$, i.e. the $H$-nearest $\Zmat$ to $\xvect_i$.
				\STATE \alglinelabel{alg: approximate posterior} Compute $w\left(\xvect_i\right)$ using $\Zmat^H_{\xvect_i}$ as in \eqref{eq: binary form of w(xn)}.
				\STATE \alglinelabel{alg: extract relevant parameters} Extract $\mvect_{w\left(\xvect_i\right)}$ and $\Smat_{w\left(\xvect_i\right)}$ from $\mvect$ and $\Lmat$.
				\STATE \alglinelabel{alg: compute approximate predictive distribution} Compute $q\left(f_i\left|w\left(\xvect_i\right)\right.\right)$ as in \eqref{eq: compute posterior using H-nearest IP}.
				\STATE \alglinelabel{alg: update ell} $\ELL \gets \ELL + \E_{q\left(f_i\left|w\left(\xvect_i\right)\right.\right)}\log p\left(y_i\left|f_i\right.\right)$.
				\STATE \alglinelabel{alg: update kl} $\KL \gets \KL + \KL\left(q\left(\uvect_{w\left(\xvect_i\right)}\right) \| p\left(\uvect_{w\left(\xvect_i\right)}\right)\right)$
			\ENDFOR
			\STATE \alglinelabel{alg: update ll} $\tilde{\LL}_\elbo \gets \frac{N}{n_{\Brm}}\ELL - \frac{1}{n_{\Brm}}\KL$.
			\STATE \alglinelabel{alg: update parameters} Update $\thetavect$ using the derivative of $\tilde{\LL}_\elbo$.
		\ENDWHILE
	\end{algorithmic}
\end{algorithm}

\paragraph{Predictive distribution}

Contrary to what happens at training time, where mini-batches of data are drawn randomly, at test time the inputs of interest are not random; we need to describe the predictive distribution in terms of the deterministic function $w\left(\xvect\right)$.
In fact, if we would like to approximate the predictive distribution at $\xvect_n$ using $H$-nearest inducing inputs to $\xvect$, i.e. $\Zmat^H_{\xvect_n}$, then $w\left(\xvect\right) = \left[w^{(1)}_\xvect ... w^{(M)}_\xvect\right]^T$ where,
\begin{equation}
\label{eq: binary form of w(xn)}
w^{(m)}_\xvect = 
\begin{cases}
1 & \mbox{if } \zvect_m \in \Zmat^H_{\xvect}\\
0 & \mbox{else}
\end{cases}
\text{ , with } m = 1, ..., \nind
\end{equation}
We extract the relevant elements using $w\left(\xvect\right)$;
for the mean, we have $\mvect_{w\left(x_i\right)} = \Dmat_{w\left(x_i\right)} \mvect$, and for the covariance we select the appropriate rows and columns using $\Smat_{w\left(\xvect_i\right)} = \Dmat_{w\left(\xvect_i\right)} \Smat \Dmat_{w\left(\xvect_i\right)}$.
The approximate posterior over $f_i$ given $w\left(\xvect_i\right)$, i.e. $q\left(f_i\left|w\left(\xvect_i\right)\right.\right)$ is:
\begin{equation}
	\label{eq: compute posterior using H-nearest IP}
	\begin{split}
		\norm\bigg(f_i \mid &\Amat_{\xvect_i} \mvect_{w\left(\xvect_i\right)}, \\
		&\Kmat_{\xvect_i} + \Amat_{\xvect_i}\left(\Smat_{w\left(\xvect_i\right)} - \Kmat_{\Zmat^H_{\xvect_i}}\right)\Amat_{\xvect_i}^\top\bigg) \text{,}
	\end{split}
\end{equation}
where $\Amat_{\xvect_i} = \Kmat_{\xvect_i, \Zmat_{\xvect_i}^H}\Kmat_{\Zmat_{\xvect_i}^H}^{-1}$.

\paragraph{Limitations}
What we presented allows one to compute predictive distributions for individual test points following Eq.~\eqref{eq: compute posterior using H-nearest IP}.
As a result, it is not possible to obtain a full covariance across predictions over multiple test points.
While most performance and uncertainty quantification metrics do not consider covariances, this might be a limitation in applications where covariances are essential.
One way to work around this is to consider the union of nearest neighbors at test time; however, this would be inconsistent with Eq.~\eqref{eq: compute posterior using H-nearest IP} and the training procedure should be modified accordingly considering the union of nearest neighbors for training points within mini-batches.
The considerations suggest a further limitation of \idea, that is that the predictive distribution becomes dependent on the number of test points that are consdered within a batch at test time.
Nevertheless, the model is built and trained to deal with sparsity at test time, and this kind of inconsistency appears only when testing the model in conditions which are different compared to those at training time. 
Besides, the experiments show that \idea performs extremely well against various state-of-the-art competitors on a wide range of experiments. 

\textbf{One-dimensional regression example.}
We visualize the posterior distribution for a synthetic dataset generated on a one-dimensional input space.
We execute \svgp and \idea, and depict the posterior distributions of these two methods by showing the predictive means (orange lines) and the 95\% credible intervals (shaded areas) in Figure \ref{fig: posterior svgp, swsgp}.
We consider identical settings for the two methods (i.e.\ $128$ inducing points, kernel parameters, likelihood variance) and a neighbor area of $16$ for \idea; a full account of the setup can be found in the supplement.
We see that although the models are different, the predictive distributions appear remarkably similar. 
A more extensive evaluation follows in Section \ref{sec:experiments}.


\begin{figure}[t]
	\begin{subfigure}{.98\linewidth}
		\centering
		\includegraphics[width=.98\linewidth]{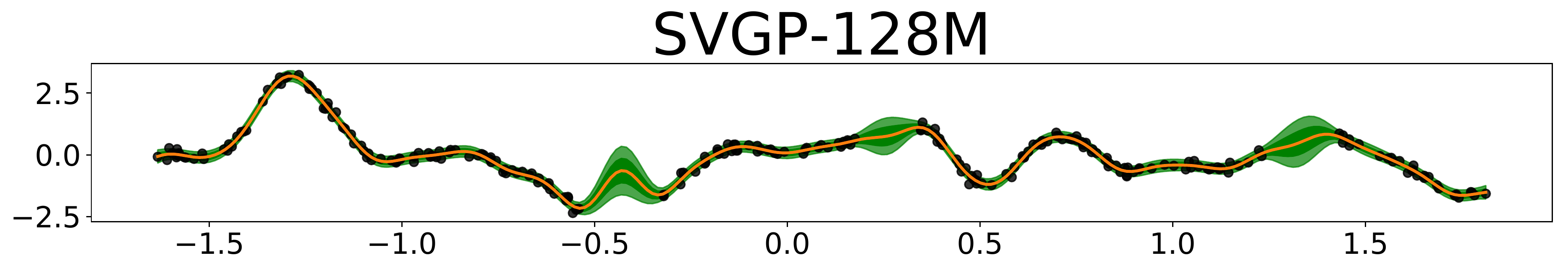}
		\label{fig: swsgp_irradiance_64_8 trainable Z}
	\end{subfigure}
	\\
	\begin{subfigure}{.98\linewidth}
		\centering
		\includegraphics[width=.98\linewidth]{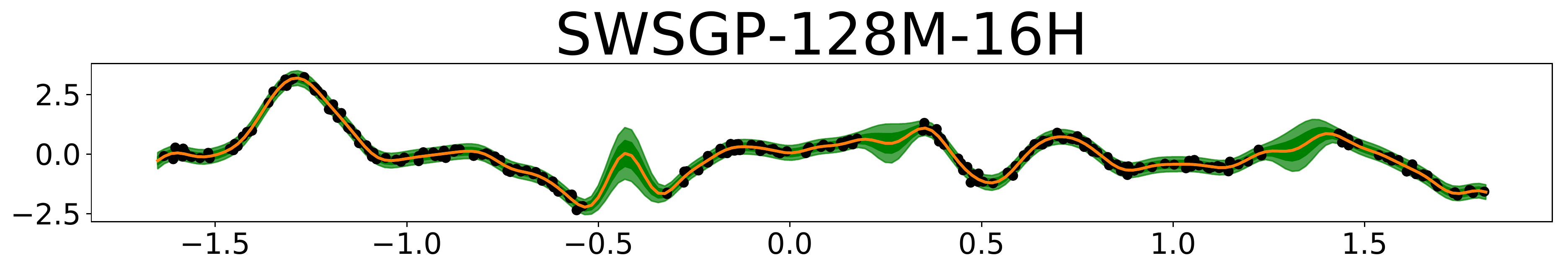}  
		\label{fig: swsgp_irradiance_64_8 fixed Z}
	\end{subfigure}
	\caption{Visualization of posterior distribution of \svgp and \idea. In both cases, we consider 128 inducing points; in terms of our scheme (\idea) we use 16 neighbors.}
	\label{fig: posterior svgp, swsgp}
\end{figure}

\subsection{Complexity}
The computational cost of \idea is dominated by lines \ref{alg: find neighbor inducing points}, \ref{alg: extract relevant parameters} and \ref{alg: compute approximate predictive distribution} in Algorithm \ref{alg: MAIP}.
For each data-point $\left(\xvect_i, y_i\right)$ in mini-batch $\Brm$, we need to find the $H$ nearest inducing neighbors $\Zmat^H_{\xvect_i}$ for $n_{\Brm}$ points in line \ref{alg: find neighbor inducing points}, where $n_{\Brm} = |\Brm|$; this contributes to the worst-case complexity by $\bigO\left(n_{\Brm} M H\right)$. 

In line~\ref{alg: extract relevant parameters}, we extract relevant parameters from $\mvect$ and $\Lmat$.
We focus on the cost of extracting $\Smat_{w\left(\xvect_i\right)}$ from $\Lmat$.
Similar to \svgp (Section \ref{subsection: svgp}), we consider $\Smat=\Lmat\Lmat^T$, where $\Lmat$ is lower triangular.
We extract $\Lmat_{w\left(\xvect_i\right)} = \Dmat_{w\left(\xvect_i\right)} \Lmat$ which contains the rows of $\Lmat$. Then, we compute $\Smat_{w\left(\xvect_i\right)}$ by $\Smat_{w\left(\xvect_i\right)} = \Lmat_{w\left(\xvect_i\right)} \Lmat_{w\left(\xvect_i\right)}^T$. 
The computational complexity of selecting the variational parameters is $\bigO\left(n_{\Brm} M H^2\right)$.

Finally, the computation of approximating the predictive distribution in line \ref{alg: compute approximate predictive distribution} requires $\bigO\left(n_{\Brm} H^3\right)$.
The overall complexity for \idea in the general case is $\bigO\left(n_{\Brm} M H + n_{\Brm} M H^2 + n_{\Brm} H^3\right)$, which is a significant improvement over the $\bigO\left(M^3\right)$ complexity of standard \svgp, assuming that $n_{\Brm}, H \ll M$.
If we choose $\Smat$ to be diagonal, the total complexity reduces to $\bigO\left(n_{\Brm} M H + n_{\Brm} H^3\right)$; if we additionally consider $\Zmat$ to be fixed, the computational cost is $O(n_{\Brm} H^3)$.
In the experiments of Section \ref{sec:experiments} we explore all these settings.

%% file: experiments.tex
\section{Experiments}
\label{sec:experiments}
We conduct experiments to evaluate \idea on a variety of experimental conditions.
Our approach is denoted by \idea-\Mip-\Hip, where $M$ inducing points are used and $H$ determines how many neighbors are selected.
The locations of the inducing points are optimized, unless stated otherwise.
We introduce \svgp-\Mip and \svgp-\Mip-\Hip as competitors; \svgp-\Mip uses $M$ inducing points.
\svgp-\Mip-\Hip, instead, refers to \svgp using $M$ inducing points at training time and $H$-nearest inducing inputs at test time.
We also use \svgp-{\footnotesize{16}\Mip} using $16$ inducing points as a reference. 
The comparison is carried out on some UCI data sets for regression and classification, i.e., \powerplant, \kin, \protein, \eeg, \credit and \spam. 
We also consider larger scale data sets, such as, \wave, \query and the \airline data or images classification on \mnist.
The task for the \airline data set is the classification of whether flights are to subject to a delay, and we follow the same setup as in \citet{Hensman13} and \citet{Wilson16}.
Regarding \mnist, we consider the binary classification setup of separating odd and even digits (pixels normalized between $0$ and $1$), and we use a Bernoulli likelihood with probit inverse link function.
We use the Mat\'{e}rn-$\nicefrac{5}{2}$ kernel in all cases except for the \airline dataset, where the sum of a Mat\'{e}rn-$\nicefrac{3}{2}$ and a linear kernel is used, similar to \citet{Hensman15b}.
All models are trained using the Adam optimizer \citep{Kingma14b} with a learning rate of $0.001$ and a mini-batch size of $64$.
The likelihood for regression and binary classification are set to Gaussian and probit function, respectively.
In regression tasks, we report the test root mean squared error (\rmse) and the test mean negative log-likelihood (\mnll), whereas we report the test error rate (\err) and \mnll in classification tasks.
The results are averaged over $10$ folds.
All models in section \ref{sec: listM-listH} are trained over $300,000$ iterations.
\input{increasing_M_H}
\input{runtime}
\input{large_M}

\input{comparing_ski}
\input{comparing_localgp}
\input{joint_predictive_covariances}

%% file: increasing_M_H.tex
\subsection{Impact of M and H}
\label{sec: listM-listH}
\begin{figure}[!htb]
	\centering
	\begin{subfigure}{.9\linewidth}
		\centering
		\includegraphics[width=.98\linewidth]{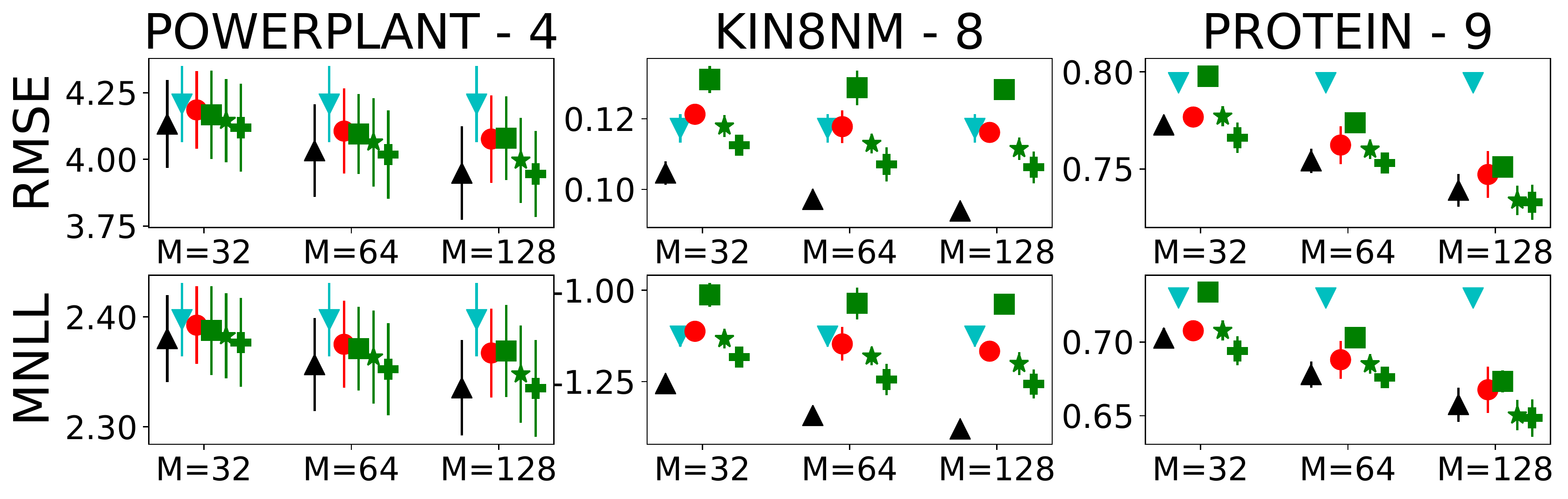}
		\label{fig: listM-listH-regression}
	\end{subfigure}
	\begin{subfigure}{.9\linewidth}
		\centering
		\includegraphics[width=.98\linewidth]{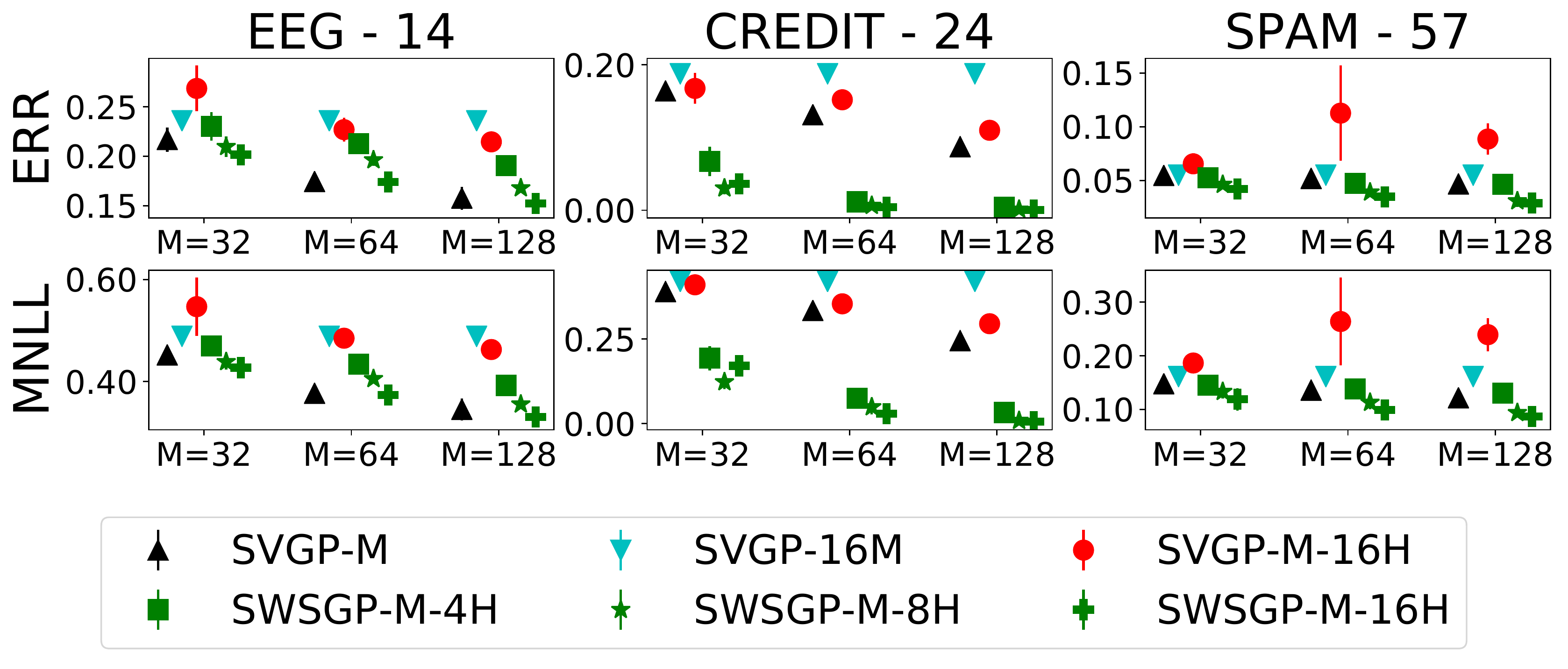}
		\label{fig: listM-listH-binary-classification}
	\end{subfigure}
	\caption{
		Evaluation of \idea on \uci data sets with various configuration for $M$ and $H$. 
		The title of each sub-figure follows the format of [name of data set]-[data dimensions].
		The black up-triangles are for \svgp with $M$ inducing points.
		The cyan down-triangles are for \svgp with $16$ inducing points.
		The red circles are for \svgp training with $M$ inducing points and the prediction at an unseen data $\xvect$ are made by $\Zmat_\xvect^H$.
		The green squares, stars and plus are for \idea with $H$ of $4$, $8$ and $16$ respectively.
		In these experiments, $M$ varies from $32$ to $128$, as shown on horizontal axes.
		The standard deviation of the error metrics over 10 folds is represented by vertical bars; they are very small for most configurations. 
	}
	\label{fig: listM-listH}
\end{figure}

We begin our evaluation by investigating the behavior of \swsgp-\Mip-\Hip with respect to $M$ and $H$.
Figure \ref{fig: listM-listH} shows that \swsgp-\Mip-\Hip consistently outperforms \svgp-\Mip-\Hip.
This suggests that including neighbor information at prediction time, combined with the use of a larger set of inducing points alone is not enough to obtain competitive performance, and that only thanks to the sparsity-inducing prior over latent variables, this yields improvements.
Crucially, the performance metrics obtained by \swsgp are comparable with those obtained by \svgp-\Mip, while at each iteration only a small subset of $H$ out of $M$ inducing points are updated, carrying a significant complexity reduction.

%% file: runtime.tex
\subsection{Running Time}
\begin{table}[!htb]
	\centering
    \caption{Comparison of running time between \svgp and \idea.
    	The running time for a training iteration is denoted by $t_1$.
    	The testing time for an unseen example is denoted by $t_2$.
    	Times are in milliseconds.
    	The corresponding evaluated metrics ,i.e. \err and \mnll are also shown in terms of training time.
    	These figures are averaged over 5 folds.
    	The models for \eeg and \mnist are trained in 90 minutes and 6 hours respectively.
    	\vspace{5pt}
    }
	\label{tab: running time}
	\begin{subtable}[tb]{0.98\linewidth}
		\centering
		\caption{Running times for \svgp-{\scriptsize{256}}\Mip and \idea-{\scriptsize{256}}\Mip-{\scriptsize{4}}\Hip on \eeg.}
		\begin{tabular}{C{2.4cm}C{0.9cm}C{0.9cm}C{0.9cm}C{0.9cm}}
			\toprule
			Configuration & $t_1$(ms) & $t_2$(ms) & \err & \mnll\\
			\midrule
			\svgp-{\footnotesize{256}}\Mip & $\mathbf{21.42}$ & 1.43 & $\mathbf{0.17}$ & $\mathbf{0.34}$\\
			\idea-{\footnotesize{256}}\Mip-{\footnotesize{4}}\Hip & 26.18 & $\mathbf{0.56}$ & 0.18 & 0.36\\
			\bottomrule
		\end{tabular}
	\end{subtable}
	\begin{subtable}[tb]{0.98\linewidth}
		\centering
		\vspace{5pt}
		\caption{Running times for \svgp-{\scriptsize{1024}}\Mip and \idea-{\scriptsize{1024}}\Mip-{\scriptsize{4}}\Hip on \mnist.}
		\begin{tabular}{C{2.6cm}C{0.9cm}C{0.9cm}C{0.9cm}C{0.9cm}}
			\toprule
			Configuration & $t_1$(ms) & $t_2$(ms) & \err & \mnll\\
			\midrule
			\svgp-{\footnotesize{1024}}\Mip & 516 & 21.6 & 0.02 & 0.066\\
			\idea-{\footnotesize{1024}}\Mip-{\footnotesize{4}}\Hip & $\mathbf{233}$ & $\mathbf{1.77}$ & $\mathbf{0.016}$ & $\mathbf{0.05}$\\
			\bottomrule
		\end{tabular}
	\end{subtable}
	\newline
	\begin{subtable}[tb]{0.9\linewidth}
		\centering
		\vspace{10pt}
		\includegraphics[width=.98\linewidth]{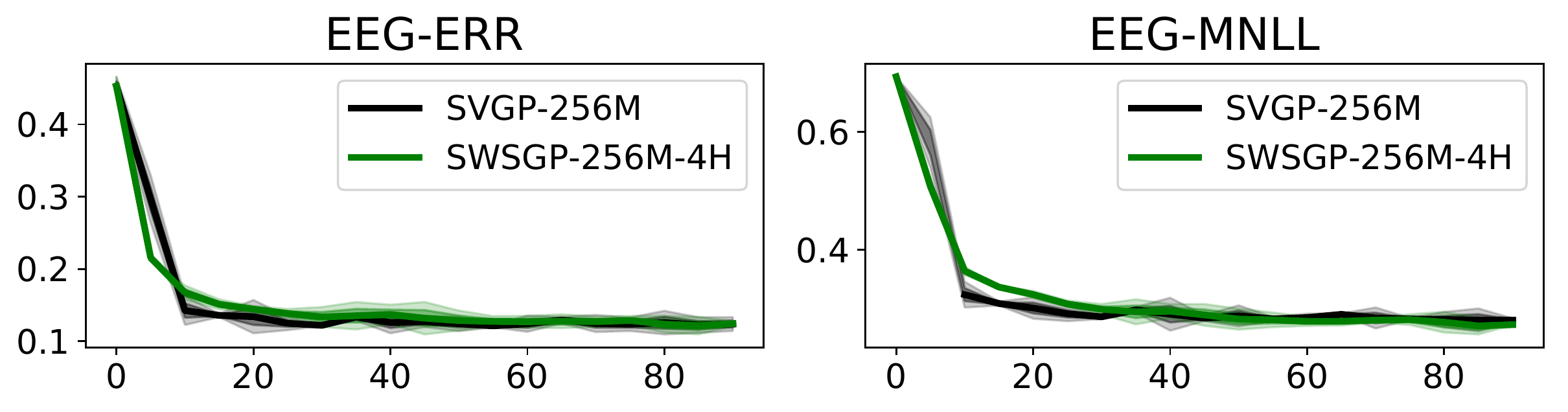}
		\caption{\err and \mnll over training time (in minutes) on \eeg.}
	\end{subtable}
    \begin{subtable}[tb]{0.9\linewidth}
    	\centering
    	\includegraphics[width=.98\linewidth]{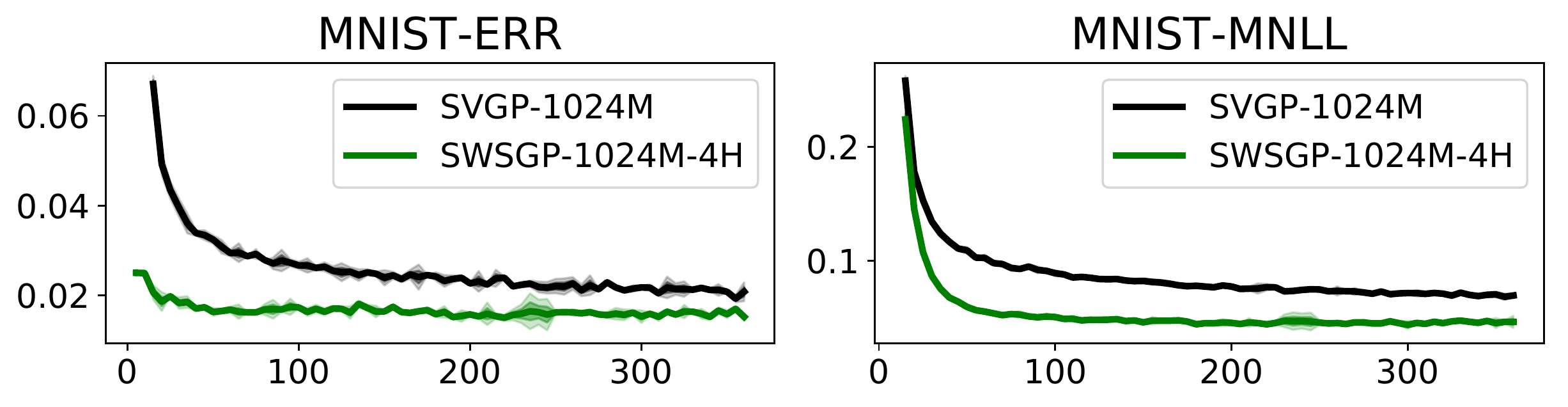}
    	\caption{\err and \mnll over training time (in minutes) on \mnist.}
    \end{subtable}
\end{table}

In Table \ref{tab: running time}, we report the training and testing times of \idea and \svgp 
In \svgp, we set $M=256$ for \eeg, and $1024$ for \mnist, i.e. \svgp-{\footnotesize{256}}\Mip and \svgp-{\footnotesize{1024}}\Mip.
In our approach, we use the same $M$ and we set $H$ to $4$, i.e. \idea-{\footnotesize{256}}\Mip-{\footnotesize{4}}\Hip and \idea-{\footnotesize{1024}}\Mip-{\footnotesize{4}}\Hip.
In Table \ref{tab: running time}, we stress that $t_1$ and $t_2$ in \idea take into account the computation of finding neighbors inducing inputs for each data-point.
In \svgp, we assume that $\Kmat_{\Zmat}^{-1}$ is pre-computed and saved after the training phase.
Therefore, the computational cost to evaluate the predictive distribution on a single test point is $\bigO\left(M^2\right)$. 
The time $t_2$ in \svgp refers to the execution time of carrying out predictions with the complexity of $\bigO\left(M^2\right)$.

The results in  Tab \ref{tab: running time} show a consistent improvement at test time compared to \svgp across all values of $H$ and $M$. 
At training time, the results show a trend dependent on the number $M$ of inducing points. 
Not surprisingly, \idea offers limited improvements when $M$ is small. 
Considering \eeg in which $M$ is set to $256$, \svgp is faster than \idea in terms of training time. 
This is because the inversion of a $256 \times 256$ matrix requires less time than finding the neighbors and inverting several $4 \times 4$ matrices.
However, Tab \ref{tab: running time} shows dramatic speedups compared to \svgp when the number of inducing points $M$ is large.
When $M=1024$ on \mnist, \idea-{\footnotesize{1024}}\Mip-{\footnotesize{4}}\Hip is faster than \svgp-{\footnotesize{1024}}\Mip at training time. 
This is due to the inversion of the $1024 \times 1024$ kernel matrix being a burden for \svgp, whereas \idea deals with much cheaper computations.
In addition, we show the corresponding \err and \mnll of each model when we train \svgp and \idea on \eeg and \mnist.
On the \eeg data set, our method is comparable with \svgp.
On \mnist, 
\idea reaches high accuracies significantly faster compared to \svgp.



%% file: large_M.tex
\subsection{Large-scale Sparse GP Modeling with a Huge Number of Inducing Points}
Here we show that \idea allows one to use sparse \gps with a massive number of inducing points without incurring a prohibitive computational cost.
We employ several large-scale data sets, i.e. \protein, \eeg, \wave, \query and \airline with 5 millions training samples.
We test \idea with a large number of inducing points, ranging from $3,000$ to $100,000$.
In this case, we keep the inducing locations fixed for \idea.
We have attempted to run \svgp with such large values of $M$ without success (out of memory in a system with 32GB of RAM). 
Therefore, as a baseline we execute \svgp-{\footnotesize{128}}\Mip and \svgp-{\footnotesize{512}}\Mip and report the results of \svgp with the configuration in \citet{Hensman15b}.

In \idea, we impose a diagonal matrix $\Smat$ in the variational distribution $q\left(\uvect \mid \wvect\right)$, 
and we fix the position of the inducing inputs during training.
By fixing the inducing inputs, we can operate with pre-computed information about which inducing inputs are neighbors of training inputs.
Thanks to these settings, \idea's training phase requires $\bigO\left(n_{\Brm} H^3\right)$ operations only, where $n_{\Brm}$ is the mini-batch size.
Due to the appropriate choice of $H$ and $n_{\Brm}$, and the computational cost being independent of $M$, unlike \svgp, we can successfully run \idea with an unprecedented number of inducing point, e.g. $M = 100,000$. 

\begin{figure}[!htb]
	\centering
	\begin{subfigure}{.9\linewidth}
		\centering
		\includegraphics[width=.98\linewidth]{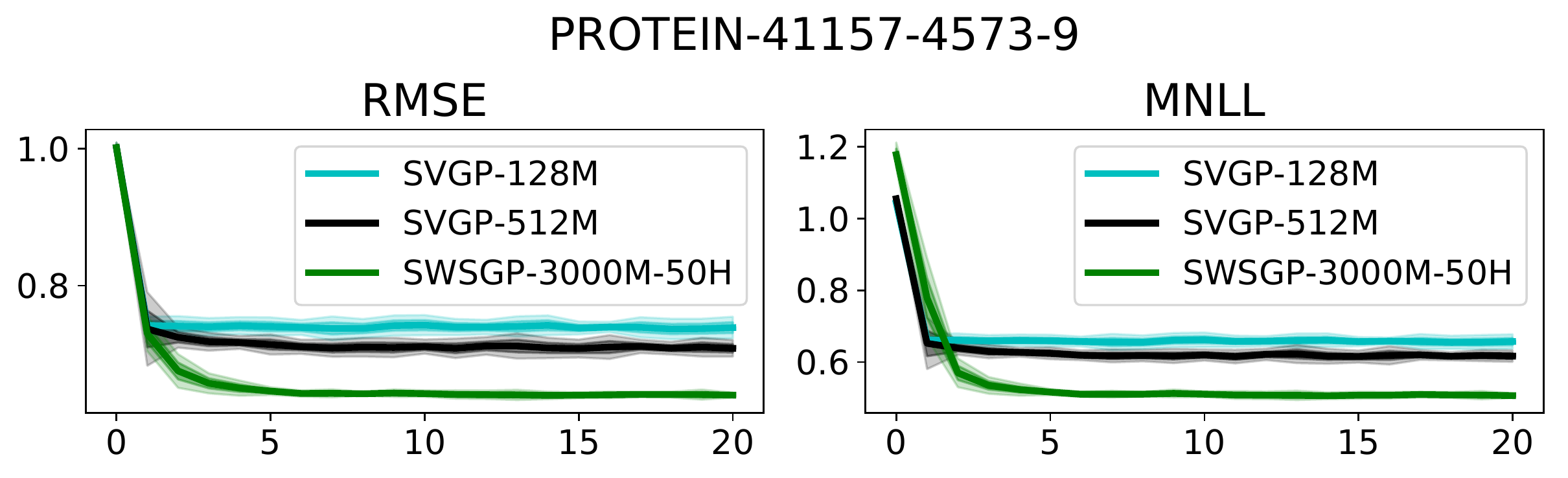}
		\label{fig: protein runtime}
	\end{subfigure}
    \begin{subfigure}{.9\linewidth}
    	\centering
    	\includegraphics[width=.98\linewidth]{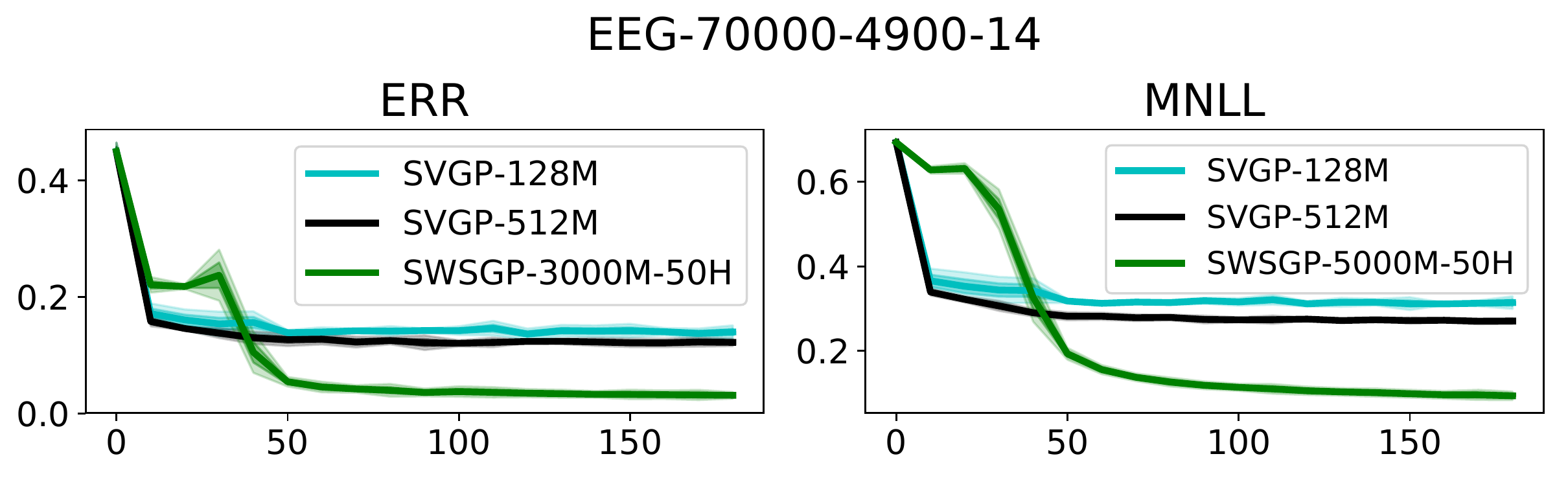}
    	\label{fig: eeg runtime}
    \end{subfigure}
	\begin{subfigure}{.9\linewidth}
		\centering
		\includegraphics[width=.98\linewidth]{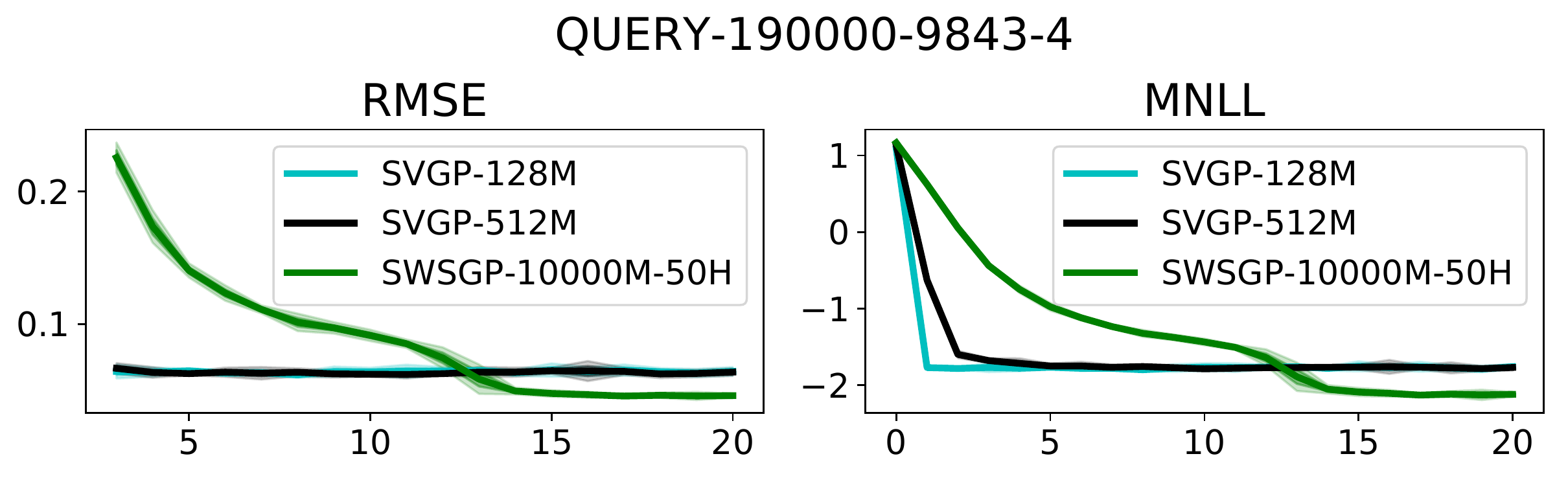}
		\label{fig: query runtime}
	\end{subfigure}
    \begin{subfigure}{.9\linewidth}
    	\centering
    	\includegraphics[width=.98\linewidth]{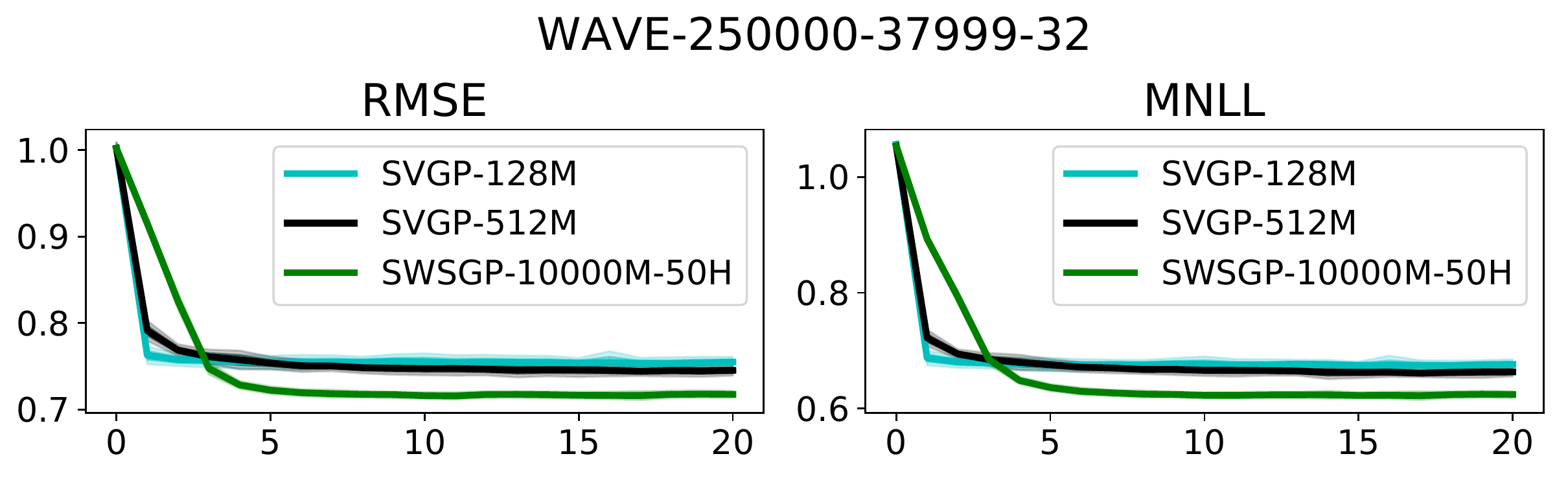}
    	\label{fig: wave runtime}
    \end{subfigure}
    \caption{\swsgp with large number of inducing points. The figure shows the progression of \rmse (\err) and \mnll over time. Horizontal axes indicate the running times in minute.
    The title of each sub-figure follows the format of [name of data set]-[training size]-[testing size]-[data dimensions].}
	\label{fig: large-M}
\end{figure}
In \airline experiments, by setting $H$ and the mini-batch size $n$ to $100$ and $64$ respectively, in about $24$ hours of training we could run \idea-{\footnotesize{100,000}}\Mip-{\footnotesize{100}}\Hip for one million iterations.
The \err and \mnll of \idea-{\footnotesize{100,000}}\Mip-{\footnotesize{100}}\Hip evaluated on the test set are $21\%$ and $0.48$, respectively, while the \err and \mnll of \svgp-{\footnotesize{200}}\Mip published in \citet{Hensman15b} are about $34\%$ and $0.61$, respectively.
To the best of our knowledge, \idea is the first to enable sparse \gps with such a large set of inducing points without imposing a grid structure on the inducing inputs.
In addition, in Figure~\ref{fig: large-M} we also show that \swsgp using $3,000$ or $5,000$ or $10,000$ inducing points outperforms \svgp using $128$ and $512$ inducing points on other data sets.

%% file: comparing_ski.tex
We conclude by reporting comparisons with other \gp-based models. 
In particular, we compare against the Stochastic Variational Deep Kernel Learning (\svdkl) \citep{Wilson16} and the Deep \gp approximated with random features (\dgprbf) \citep{Cutajar17}. 
In the former, \kissgp is trained on top of a deep neural network which is optimized during training, and in the latter the layers of a deep \gp are approximated as parametric models using random feature expansions. 
Both competitors feature mini-batch based learning, so this represents a challenging test for \idea.
The results in Table \ref{tab: comparison of swsgp and ski} show that \idea is comparable with these competitors on various data sets.
We believe that this is a remarkable result obtained by our shallow \idea, supporting the conclusions of previous works showing that advances in kernel methods can result in performance which are competitive with deep learning approaches (see, e.g., \citet{Rudi17}).
\begin{table}[!htb]
\caption{Comparison of \idea, \kissgp \citep{Wilson15}, \svdkl\citep{Wilson16} and \dgprbf \citep{Cutajar17}.
	The results \swsgp and \kissgp are averaged over 5 folds. 
In order to deal with the difficulties of \kissgp to handle large-dimensional input spaces, we followed \citep{Wilson15a}, and we linearly projected the inputs to a two-dimensional space using a linear transformation that is learned at training time, and used a grid of size 100.
}
\label{tab: comparison of swsgp and ski}
\begin{subtable}{.98\linewidth}
	\centering
	\caption{\airline}
	\label{tab: comparing ski airline}
	\begin{tabular}{C{2.9cm}C{2.0cm}C{2.0cm}}
		\toprule
		Method & \err & \mnll\\
		\midrule
		\idea-{\footnotesize{100k}}\Mip-{\footnotesize{100}}\Hip & $\mathbf{0.210 \pm 0.012}$ & $0.48 \pm 0.015$\\
		\svdkl & $0.22$ & $\mathbf{0.46}$ \\
		\dgprbf & $\mathbf{0.21}$ & $\mathbf{0.46}$ \\
		\bottomrule
	\end{tabular}
\end{subtable}
\begin{subtable}{0.98\linewidth}
	\vspace{10pt}
	\centering
	\caption{\powerplant}
	\label{tab: comparing ski powerplant}
	\begin{tabular}{C{2.3cm}C{2.1cm}C{2.1cm}}
		\toprule
		Method & \rmse & \mnll\\
		\midrule
		\idea-{\footnotesize{64}}\Mip-{\footnotesize{4}}\Hip & $\mathbf{4.095 \pm 0.145}$ & $\mathbf{2.371 \pm 0.038}$ \\
		\kissgp & $4.459 \pm 0.355$ & $3.074 \pm 0.037$\\
		\bottomrule
	\end{tabular}
\end{subtable}
\begin{subtable}{0.98\linewidth}
	\vspace{10pt}
	\centering
	\caption{\protein}
	\label{tab: comparing ski protein}
	\begin{tabular}{C{2.3cm}C{2.1cm}C{2.1cm}}
		\toprule
		Method & \rmse & \mnll\\
		\midrule
		\idea-{\footnotesize{64}}\Mip-{\footnotesize{4}}\Hip & $\mathbf{0.773 \pm 0.002}$ & $\mathbf{0.702 \pm 0.003}$\\
		\kissgp & $0.816 \pm 0.001$ & $0.904 \pm 0.017$\\
		\bottomrule
	\end{tabular}
\end{subtable}
\end{table}

%% file: comparing_localgp.tex
\subsection{Comparison to Local GPs}
We finally demonstrate that \idea behaves differently from other approaches that use local approximations of \gps.
We consider two well-established approaches of local \gps proposed by \citet{Kim05} and \citet{Urtasun08}.
Following \citet{Liu2020WhenGP}, we refer to these methods as \emph{Inductive} \gps and \emph{Transductive} \gps, respectively.
We run all methods on two regression data sets: \powerplant and \kin.
We set the number of local experts to $64$, and we use the same number of inducing points for \idea (with $H$ either 4 or 8).
As the size of \powerplant and \kin are approximately $7,000$, we set the number of training points governed by a local expert to $100$.
For the local \gp approaches, we choose $64$ locations in the input space using the $K$-means algorithm, and for each location we choose $100$ neighboring points; we then train the corresponding local \gp expert.
For the testing phase, inductive \gps simply rely on the nearest local experts to an unseen point $\xvect_*$.
For transductive \gps, we use $100$ neighbors of $\xvect_*$ and the nearest local expert to make predictions.
In Table \ref{tab: Comparison to Local approximation of gps}, we summarize \rmse and \mnll for all methods; \idea clearly outperforms the local \gp approaches in terms of \mnll.

\begin{table}[!htb]
	\centering
    \caption{Comparison with Local \gp approximations.}
    \vspace{8pt}
    \label{tab: Comparison to Local approximation of gps}
	\begin{tabular}{C{2.8cm}C{2.0cm}C{2.0cm}}
	\toprule
	Method & \powerplant & \kin\\
	       & \rmse $\mid$ \mnll & \rmse $\mid$ \mnll\\
	\midrule
	\idea-64-4 & $4.27 \mid 2.41$ & $0.11 \mid -1.27$\\
	\midrule
	\idea-64-8 & $\textbf{4.24} \mid \textbf{2.40}$ & $0.10 \mid \textbf{-1.38}$\\
	\midrule
	Inductive GPs & $9.93 \mid 38.38$ & $0.13 \mid -0.40$\\
	\midrule
	Transductive GPs & $6.17 \mid 18.78$ & $\textbf{0.09} \mid -0.65$\\
	\bottomrule
	\end{tabular}
\end{table}

%% file: joint_predictive_covariances.tex
\subsection{Joint Predictive Covariances}
\begin{figure}[!htb]
	\centering
	\includegraphics[width=.98\linewidth]{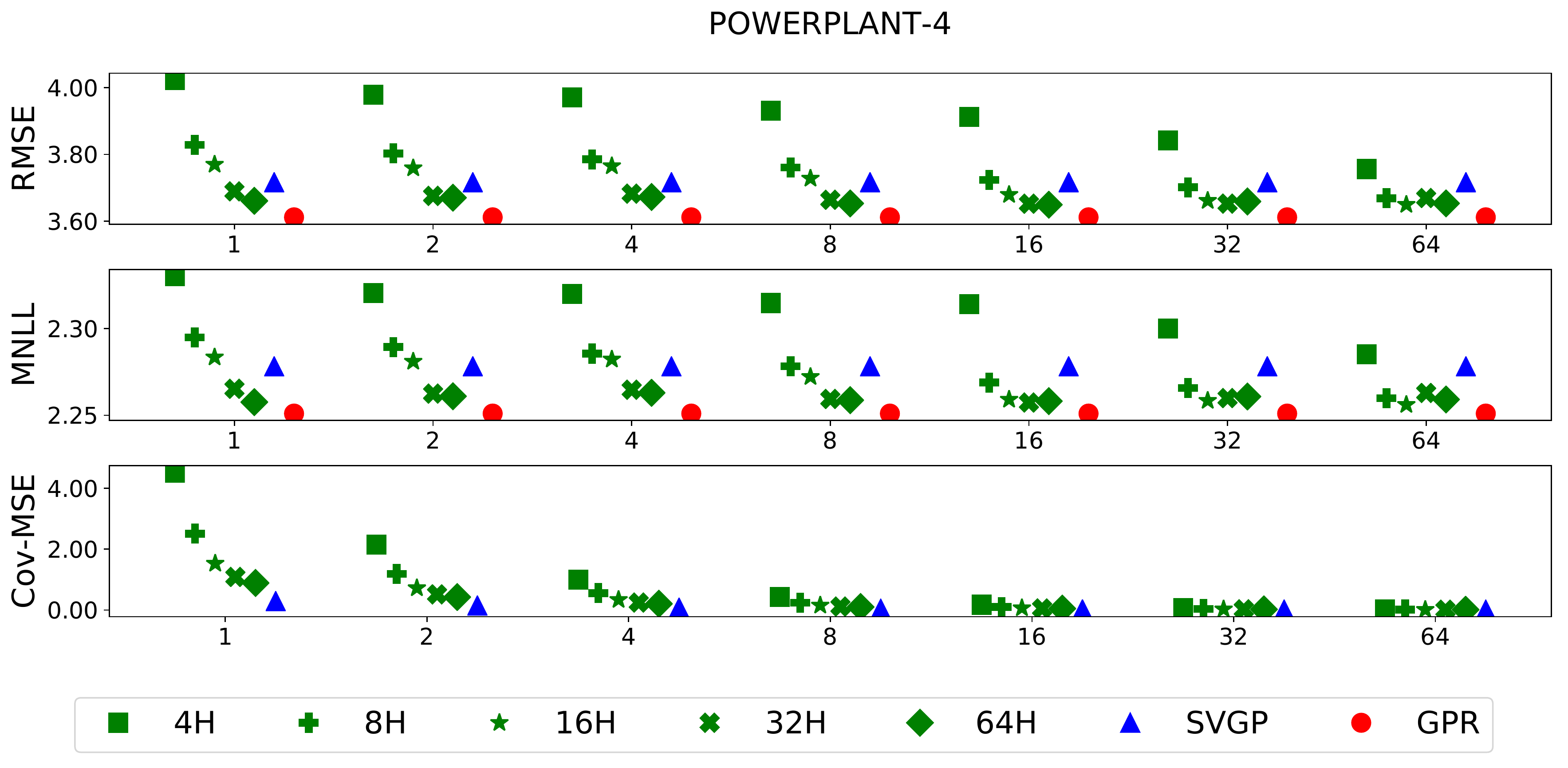}
	\caption
	{
		\swsgpu on \powerplant.
		The horizontal axis shows the mini-batch size of the testing set.
		We evaluate \swsgpu with various numbers of nearest inducing points from $4$ to $64$, i.e. $4H$, $8H$, $16H$, $32H$, and $64H$.
		The green markers show \swsgpu.
		For example, in order to plot the green squares $4H$, we firstly find the 4-nearest inducing points of each sample within a testing batch.
		Then, we use the union of these nearest inducing points to make a joint prediction for all samples in the testing batch.
	}
	\label{fig: swsgpu on powerplant}
\end{figure}
In this section, we propose an extension of \idea where the training procedure is modified by considering the union of nearest neighbors for training points within mini-batches.
We refer to this as $\swsgpu$, where \name{u} stands for ``union''.
In the experiments, we execute \swsgpu-{\footnotesize{512}}\Mip-{\footnotesize{128}}m where the total number of inducing points is $512$, and the number of active inducing points in each training iteration is fixed to 128.
To illustrate the effectiveness of the modification, we compare \swsgpu-{\footnotesize{512}}\Mip-{\footnotesize{128}}m against \svgp-{\footnotesize{512}}\Mip and full \gps.
Due to the computational complexity of full \gps, we only use $2,000$ training samples and fix the computational budget to six hours.
In the evaluation phase, we randomly group the test set based on the specified mini-batch size.
Next, the joint predictive distributions of each testing batch are obtained, which we use to compute \rmse and \mnll.
The results in Figure~\ref{fig: swsgpu on powerplant} are averaged by repeating the process $10$ times.
The results indicate that \swsgpu works well when using an adequate number of nearest inducing points.
In the figure, we also assess the quality of the covariance approximation of  \swsgpu and \svgp by reporting the mean square error with respect to the covariance of full \gps. 
These results show that \swsgpu provides a good approximation of joint predictive covariances.

%% file: conclusions.tex
\section{Conclusions}
Sparse approaches that rely on inducing points 
have met with success in reducing the complexity of \gp regression and classification.
However, these methods are limited by the number of inducing inputs that is required to obtain an accurate approximation of the true \gp model.
A large number of inducing inputs is often necessary in cases of very large datasets, which marks the limits of practical applications for most \gp-based approaches.

In this work, we further improve the computational gains of sparse \gps.
We proposed \idea, a novel methodology that imposes a hierarchical and sparsity-inducing effect on the prior over the inducing variables.
This has been realized as a conditional \gp given a random subset of the inducing points, which is defined as the nearest neighbors of random mini-batches of data.
We have developed an appropriate variational bound which can be estimated in an unbiased way by means of mini-batches. 
We have performed an extensive experimental campaign that demonstrated the superior scalability properties of \idea compared to the state-of-the-art.